\documentclass{article}

\PassOptionsToPackage{numbers,sort&compress}{natbib}

\usepackage[preprint]{neurips_2026}

\usepackage[utf8]{inputenc} 
\usepackage[T1]{fontenc}    
\usepackage{hyperref}       
\usepackage{url}            
\usepackage{booktabs}       
\usepackage{amsfonts}       
\usepackage{amsmath}        
\usepackage{amssymb}        
\usepackage{subcaption}     
\usepackage{nicefrac}       
\usepackage{microtype}      
\usepackage{xcolor}         
\usepackage{graphicx}       
\usepackage{multirow}       
\usepackage{array}          
\usepackage{wrapfig}

\newcommand{\cmark}{\textcolor{green!55!black}{\checkmark}}
\newcommand{\xmark}{\textcolor{red!70!black}{\boldmath$\times$\unboldmath}}
\newcommand{\pmark}{\textcolor{orange!85!black}{$\blacktriangle$}}
\newcommand{\corrauth}{\textsuperscript{\rm \dag}}

\title{Rollout Pass-Rate Control: Steering Binary-Reward RL Toward Its Most Informative Regime}

\author{%
  \normalfont
  Tianshu Zhu$^{1}$, Wenyu Zhang$^{1}$, Xiaoying Zuo$^{1}$, Lun Tian$^{1}$, Haotian Zhao$^{1}$, \\
  Yucheng Zeng, Jingnan Gu$^{1}$\corrauth, Daxiang Dong$^{1}$\corrauth, Jianmin Wu$^{1}$\corrauth, \\
  Dawei Yin$^{1}$\corrauth, Dou Shen$^{1}$\corrauth \\
  $^{1}$Baidu \\
  Emails: \texttt{\{zhutianshu,zhangwenyu08,zuoxiaoying,tianlun,zhaohaotian02,} \\
  \texttt{gujingnan,dongdaxiang,wujianmin,yindawei02,shendou\}@baidu.com} \\
  \corrauth Corresponding authors: Jingnan Gu, Daxiang Dong, Jianmin Wu, Dawei Yin, and Dou Shen
}

\begin{document}

\maketitle

\begin{abstract}
Agentic reinforcement learning (RL) for software engineering spends much of its compute on stateful trajectories whose grouped binary rewards are highly skewed and weakly contrastive. We frame this as pass-rate control and show that the binary reward-side signal is strongest near a 50\% rollout pass rate under four criteria: reward entropy, group-filtering survival, leave-one-out (RLOO) advantage energy under Group Relative Policy Optimization (GRPO), and success--failure pair count. We propose \emph{Prefix Sampling} (PS), which replays self-generated trajectory prefixes to steer skewed groups toward this regime: successful prefixes give mostly failing groups a head start, while failing prefixes handicap mostly passing groups. Replayed states are reconstructed through the existing rollout path, and replayed tokens are masked from the loss so optimization applies only to current-policy continuations. On SWE-bench Verified, PS reaches the baseline high-score regime within evaluation variability while delivering $2.01\times$/$1.55\times$ end-to-end wall-clock speedups on Qwen3-14B/32B; the 14B peak improves from $0.274$ to $0.295$. AIME 2025 experiments on 4B/8B show the same pass-rate-control pattern, and 4B ablations attribute gains to replay, bidirectional coverage, and adaptive control.
\end{abstract}

\section{Introduction}

Reinforcement learning with verifiable rewards (RLVR) is now widely used for post-training on tasks with clear external feedback, including software engineering and mathematical reasoning \citep{shao2024deepseekmath,yu2025dapo,deepswe2025}. Within this paradigm, online policy-gradient methods such as Group Relative Policy Optimization (GRPO) rely on rollout trajectories to produce contrastive learning signal \citep{shao2024deepseekmath}. As these tasks move toward agentic reinforcement learning, rollouts become longer, more interactive, and substantially more expensive to generate \citep{deepswe2025,letitflow2025}. This raises a basic efficiency question: \textbf{when rollouts dominate compute and wall-clock time, how much of that effort actually produces useful learning signal?} Existing filtering heuristics show that not all rollouts are equally valuable, but the relationship among rollout pass rates, compute, and training benefit remains under-specified.

To reduce compute spent on uninformative rollouts, RLVR systems often filter reward-degenerate groups. We use \emph{group filtering} to denote the GRPO-style practice of discarding rollout groups whose binary rewards are all zero or all one. This operation is sometimes called rejection sampling in RLVR implementations, but differs from positive-only rejection sampling, which keeps only successful samples for supervised fine-tuning (SFT) or reinforcement. The need for such filtering is implicit in DeepSeekMath/GRPO, since group-relative normalization gives no relative advantage signal when every reward in a group is identical \citep{shao2024deepseekmath}. This exclusion is later formalized as dynamic sampling, replacing all-zero/all-one groups until the batch is filled \citep{yu2025dapo}. Beyond filtering, prior work uses prefixes or hints \citep{prefixrl2025,pope2025,stephint2025,adhint2025,hintgrpo2025,selfhinting2025}, rollout/batch reallocation \citep{yu2025dapo,dynamo2025}, SFT/off-policy mixtures \citep{luffy2025,srft2025,relift2025,uft2025}, or difficulty calibration \citep{seele2025}. These methods address difficulty or allocation in complementary ways, but do not directly steer retained skewed groups through bidirectional prefix replay under a fixed rollout budget.

We therefore formulate Rollout Pass-Rate Control (Rollout-PRC) for binary-reward RLVR tasks such as software engineering and mathematical reasoning \citep{jimenez2024swebench,hendrycks2021measuring}. For a task with multiple rollouts, the rollout pass rate is the fraction of successful rollouts in the group. In our training traces, even after group filtering, many retained rollout groups in SWE-bench-style tasks remain highly skewed, such as $1/8$ or $7/8$, and provide weak contrast. Four complementary quantities---reward entropy, group-filtering survival, leave-one-out (RLOO) advantage energy under GRPO, and success--failure pair count---all identify the same target: \textbf{training is most informative when rollout pass rates are close to 50\%} \citep{cover2006information,shao2024deepseekmath,williams1992rloo}. Rollout-PRC turns pass-rate management from an implicit byproduct of training into an explicit control objective.

Guided by Rollout-PRC, we propose Prefix Sampling (PS), a bidirectional framework that routes skewed rollout groups toward the 50\% regime through trajectory prefix replay. For mostly failing groups, it replays a successful prefix as a head start; for mostly passing groups, it replays a failing prefix as a handicap. Unlike prior prefix/hint methods that often require oracle/reference prefixes, teacher traces, hint calibration, or a separate self-hint path \citep{prefixrl2025,pope2025,stephint2025,adhint2025,hintgrpo2025,selfhinting2025}, Prefix Sampling reuses trajectories produced by the current policy itself. In stateful agent environments, it reconstructs prefix state through the existing rollout path and masks replayed tokens from the loss, so optimization is restricted to fresh continuations generated by the current policy. On SWE-bench-style agentic RL, it reaches the baseline's best verified-score regime in up to $1.76\times$ fewer evaluation steps and delivers $2.01\times$ and $1.55\times$ end-to-end speedups on Qwen3-14B \citep{qwen3_14b} and Qwen3-32B \citep{qwen3_32b}.

Our contributions are:
\begin{itemize}
	\item We identify Rollout Pass-Rate Control (Rollout-PRC) as a principled efficiency objective for grouped binary-reward RLVR updates, and show that a 50\% rollout pass rate maximizes reward entropy, group-filtering survival, RLOO advantage energy under GRPO, and success--failure pair count.

	\item We propose Prefix Sampling, a bidirectional pass-rate steering framework that replays trajectory prefixes to move skewed rollout groups toward this high-signal regime. It supports stateful agents by reconstructing prefix state through replay-through-execution, masking replayed tokens, and adapting prefix length by pass-count bucket.

	\item We demonstrate cross-scale gains on SWE-bench-style agentic RL and AIME 2025 reasoning \citep{aime2025}, including up to $1.76\times$ fewer benchmark-evaluation steps, $2.01\times$ end-to-end speedup, and mechanistic and ablation evidence showing that the gains stem from steering rerollout pass rates toward the 50\% target.
\end{itemize}

\section{Method}
\label{sec:method}

\begin{figure}[!t]
	\centering
	\includegraphics[width=0.95\linewidth]{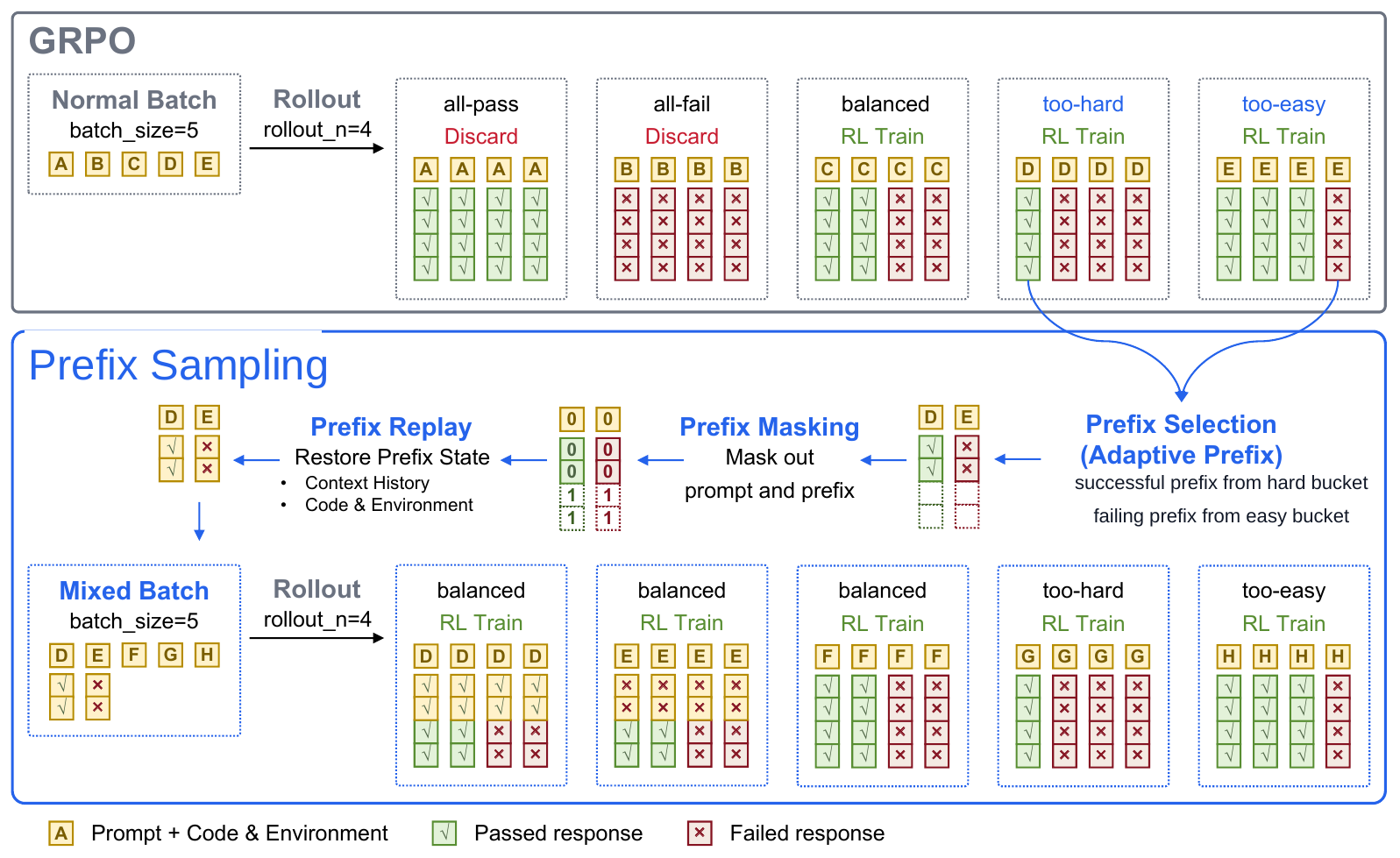}
		\caption{\textbf{Prefix Sampling pipeline.} For each task we sample a rollout group and route it by pass count: degenerate $0/8$ or $8/8$ groups are filtered; all other groups remain trainable; skewed non-degenerate groups additionally provide replay prefixes. Mostly failing hard buckets reuse a successful prefix as a head start, while mostly passing easy buckets reuse a failing prefix as a handicap. The current policy generates fresh continuations from the replay-reconstructed prefix state; masking applies RL loss only to continuation tokens, steering rerollouts toward $50\%$ without crediting replayed actions. Counts in the diagram are schematic; experiments use $N=8$ rollouts.}
	\label{fig:main}
\end{figure}

\subsection{Why target 50\% rollout pass rate}
\label{sec:why-target-50}
Given the limitations of implicit filtering, a natural question is which rollout pass rate yields the strongest binary-reward learning signal. For $N$ rollouts with pass probability $p$, random success count $K$, and observed success count $k$, Equation~\ref{eq:binary-signal-target} summarizes four complementary quantities:
\begin{equation}
\label{eq:binary-signal-target}
\left\{
\begin{aligned}
H(p) &= -p\log_2 p-(1-p)\log_2(1-p),
& \arg\max_{p} H(p) &= 0.5,\\
S_N(p) &= \Pr(0<K<N)=1-p^N-(1-p)^N,
& \arg\max_{p} S_N(p) &= 0.5,\\
E_{\mathrm{RLOO}}(k) &= \frac{k(N-k)}{(N-1)^2},
& \arg\max_{k} E_{\mathrm{RLOO}}(k) &= N/2,\\
C(k) &= k(N-k),
& \arg\max_{k} C(k) &= N/2.
\end{aligned}
\right.
\end{equation}
The first two are maximized at $p=0.5$, while the latter two are maximized at $k=N/2$ for an observed group \citep{cover2006information,shao2024deepseekmath,williams1992rloo}. Appendix~\ref{app:optimality} gives the full derivation and finite-$N$ contrastive-pair counts. Prefix Sampling therefore treats $50\%$ pass rate as a reward-side signal target, not a global curriculum optimum, and uses saved prefixes to move skewed partially successful groups toward balanced, high-information continuations.

\subsection{Prefix selection and adaptive prefix control}
\label{sec:select-control}
For $N=8$ rollouts with pass count $k$, we discard only degenerate groups ($k \in \{0,8\}$); every non-degenerate group remains eligible for the ordinary GRPO update. Balanced groups ($k \in \{3,4,5\}$) need no replay, while skewed groups seed rerollouts: hard buckets ($k \in \{1,2\}$) save a successful trajectory as prefix, and easy buckets ($k \in \{6,7\}$) save a failing one. This is the \emph{Prefix Selection} stage in Figure~\ref{fig:main}.

The \emph{Adaptive Prefix} controller maintains one prefix ratio $r_b \in [0.05,0.95]$ per bucket $b \in \{1/8,2/8,6/8,7/8\}$. For stateful SWE agents, $r_b$ selects the first $M=\lfloor r_bT\rfloor$ environment-interaction steps of a $T$-step trajectory; for single-turn math, it selects the first $M=\lfloor r_bT\rfloor$ assistant tokens. In both cases, $M$ defines the replay boundary used by prefix masking.

For hard buckets ($k \in \{1,2\}$), larger $r_b$ gives a longer successful head start and raises pass rate; for easy buckets ($k \in \{6,7\}$), larger $r_b$ replays more of a failing trajectory and lowers pass rate. Both aim to move the rerollout pass rate toward $p=0.5$.

Because model capability drifts during training, any fixed calibration can become stale. We therefore close the loop with a per-bucket feedback controller. For each bucket $b$, the trainer maintains an exponential moving average of the rerollout pass rate $\hat{p}_b$, updated each time a replayed group from that bucket completes rerollout:
\begin{equation}
	\hat{p}_b \leftarrow (1-\alpha)\hat{p}_b + \alpha\, p_b^{(\text{new})}.
	\label{eq:ema}
\end{equation}
We use $\alpha = 0.05$ (half-life $\approx 13.5$ updates), a deadzone $\delta = 0.03$ around $0.5$, an adjustment step $\eta = 0.05$, and a five-update cooldown after each adjustment. When $\hat{p}_b$ exits the deadzone, $r_b$ shifts by $\eta$ in the direction that pulls the bucket back toward the target. The update direction reverses across the two sides: for hard buckets, $r_b$ decreases when $\hat{p}_b > 0.5 + \delta$ and increases when $\hat{p}_b < 0.5 - \delta$; for easy buckets, the rule is inverted. Thus each bucket has an independent closed-loop controller, with $r_b$ as the control knob and $\hat{p}_b$ as feedback.

\subsection{Prefix replay and prefix masking}
\label{sec:replay-mask}
\enlargethispage{3\baselineskip}
Once the controller sets replay boundary $M$, \emph{Prefix Replay} reconstructs the saved prefix state. In single-turn reasoning, this prepends selected prefix tokens. For stateful coding agents, saved responses from the selected interaction steps pass through the same action parser and environment executor as ordinary model responses, reconstructing code state, conversation history, tool outputs, and execution context without a separate sandbox-snapshot restore. From this state, the current policy generates a fresh continuation.

However, if the loss were applied to the entire trajectory, replayed prefix tokens would inherit the advantage of the new rollout and contaminate credit assignment with off-policy actions. \emph{Prefix Masking} prevents this. Let $t_{\text{cont}}$ denote the first token generated by the current policy after the replay boundary; the response mask is set to zero for all assistant tokens before $t_{\text{cont}}$. Omitting clipping, normalization, and other implementation details, the masked GRPO policy-gradient term \citep{shao2024deepseekmath,schulman2017ppo} can be written schematically as
\begin{equation}
	\mathcal{L}_{\text{GRPO}}
	= -\sum_{i} A_i \sum_{t \geq t_{\text{cont}}} \log \pi_\theta(a_{i,t} \mid s_{i,t}),
	\label{eq:masked-grpo}
\end{equation}
with $i$ indexing rollouts in the group. Replayed prefix tokens shape the context but do not contribute to the loss, so gradients originate only from current-policy actions after the replay boundary. After rerollout, non-degenerate rerollout groups are combined with all ordinary, non-replayed non-degenerate groups into the same \emph{Mixed Batch} shown in Figure~\ref{fig:main}, then optimized by the same GRPO update. Thus Prefix Sampling changes starting states while keeping continuation tokens on-policy.

\section{Experiments}
Our evaluation uses two complementary regimes. SWE-bench-style agentic RL \citep{jimenez2024swebench,deepswe2025} is the primary target: rollouts are long, stateful, and expensive, so both sample efficiency and wall-clock efficiency matter. Mathematical reasoning is a lighter single-turn setting that exposes per-task pass-count distributions and parent-to-rerollout transitions, and therefore lets us audit the mechanism more directly.

\subsection{Experimental recipe}
\label{sec:experimental-recipe}
For SWE-bench, we train Qwen3-14B \citep{qwen3_14b} and Qwen3-32B \citep{qwen3_32b} in thinking mode on R2E-Gym-Subset \citep{r2egym2025} and evaluate on SWE-bench Verified \citep{jimenez2024swebench,swebench_verified2024}. We use thinking mode for stronger long-horizon reasoning in the high-difficulty software-engineering setting \citep{qwen32025}. The baseline follows a DeepSWE-style GRPO++ backbone \citep{deepswe2025}: clip-high updates, KL-free optimization, no reward-standard-deviation normalization, length-normalized loss, leave-one-out advantages, compact filtering of invalid max-context/timeout/max-step trajectories, and zero entropy bonus. Our matched setup also applies DAPO-style group filtering for $0/8$ and $8/8$ groups \citep{yu2025dapo}.

Mathematical reasoning rollouts are single-turn, so the recipe needs no replay-through-execution state reconstruction. Our backbones are Qwen3-4B-Instruct-2507 \citep{qwen3_4b_instruct_2507} and Qwen3-8B \citep{qwen3_8b}, both run without thinking mode to model RL training on instruction-style policies \citep{qwen32025}. We train on an AceReason-Math-Subset constructed locally from AceReason-Math \citep{acereason2025}: we run Qwen3-4B-Instruct-2507 with 8 rollouts per problem, estimate each problem's empirical pass rate, and exclude problems with pass rate at least 75\%. This curation yields roughly 15,000 harder problems, and we evaluate the final models on AIME 2025 \citep{aime2025}. The baseline uses the same GRPO++ recipe but excludes components specific to agents.

Both setups share several core configuration choices. We sample $N=8$ rollouts per task. SWE-bench runs use max prompt length $4096$, while mathematical reasoning runs use max prompt length $2048$. The optimizer uses clip-high GRPO with upper clip bound $0.28$, actor learning rate $1\!\times\!10^{-6}$, and zero entropy bonus \citep{deepswe2025,shao2024deepseekmath,williams1992rloo,schulman2017ppo}. The batch size is $64$ on SWE-bench and $128$ on mathematical reasoning. Hardware is matched within each backbone: 4B uses $2\times8$ A800 GPUs, 8B uses $4\times8$ A800 GPUs, and 14B/32B use $4\times8$ B200 GPUs. In the four main runs, Prefix Sampling uses the 4-bucket adaptive controller (PS-ada). Appendix~\ref{app:config} reports the run-level settings, shared GRPO++ backbone, and PS-ada controller parameters in Tables~\ref{tab:run-config}--\ref{tab:ps-config}.

\begin{table}[!htbp]
	\centering
	\setlength{\tabcolsep}{2.5pt}
	\renewcommand{\arraystretch}{1.12}
			\caption{Main benchmark and efficiency results across four backbones. Ours is Prefix Sampling with adaptive bucket-level control (PS-ada). Pass@1 is averaged over 8 evaluation runs at the Ours peak; Baseline is measured at that same checkpoint, so score gaps are same-step gaps rather than peak-vs-peak gains. In Gain rows, the Pass@1 entry is a same-step score gap, while all efficiency/resource entries are ratios. E2E denotes end-to-end wall-clock time, and pp denotes percentage points. Timing and convergence columns report same-score-or-variability comparisons.}
	\label{tab:main-results-summary}
	\begin{tabular*}{\linewidth}{@{\extracolsep{\fill}}lllccccccc@{}}
		\toprule
			Model & Benchmark & Row & \shortstack{Same-step\\Pass@1$\uparrow$\\@ Ours peak} & \shortstack{Eval\\E2E$\downarrow$} & \shortstack{Train\\E2E$\downarrow$} & \shortstack{Eval\\step$\downarrow$} & \shortstack{Train\\step$\downarrow$} & \shortstack{Step\\time$\downarrow$} & \shortstack{Valid\\groups$\uparrow$} \\
		\midrule
			\multirow{3}{*}{Qwen3-32B} & \multirow{3}{*}{\shortstack{SWE-bench\\Verified}} & \textcolor{gray!65!black}{Baseline} & $0.368$ & $268.0$h & $258.2$h & $410$ & $395$ & $2353$s & $30.9$ \\
		\cmidrule(lr){3-10}
			& & \textbf{Ours} & \textbf{0.422} & \textbf{172.9h} & \textbf{168.1h} & \textbf{290} & \textbf{282} & \textbf{2146s} & \textbf{39.8} \\
		\cmidrule(lr){3-10}
		& & \textcolor{blue!65!black}{Gain} & \textcolor{blue!65!black}{$+5.4$ pp} & \textcolor{blue!65!black}{$1.55\times$} & \textcolor{blue!65!black}{$1.54\times$} & \textcolor{blue!65!black}{$1.41\times$} & \textcolor{blue!65!black}{$1.40\times$} & \textcolor{blue!65!black}{$1.10\times$} & \textcolor{blue!65!black}{$1.29\times$} \\
		\midrule
			\multirow{3}{*}{Qwen3-14B} & \multirow{3}{*}{\shortstack{SWE-bench\\Verified}} & \textcolor{gray!65!black}{Baseline} & $0.247$ & $133.1$h & $138.4$h & $300$ & $312$ & $1597$s & $25.9$ \\
		\cmidrule(lr){3-10}
			& & \textbf{Ours} & \textbf{0.295} & \textbf{66.3h} & \textbf{80.4h} & \textbf{170} & \textbf{206} & \textbf{1405s} & \textbf{36.7} \\
		\cmidrule(lr){3-10}
		& & \textcolor{blue!65!black}{Gain} & \textcolor{blue!65!black}{$+4.7$ pp} & \textcolor{blue!65!black}{$2.01\times$} & \textcolor{blue!65!black}{$1.72\times$} & \textcolor{blue!65!black}{$1.76\times$} & \textcolor{blue!65!black}{$1.51\times$} & \textcolor{blue!65!black}{$1.14\times$} & \textcolor{blue!65!black}{$1.41\times$} \\
		\midrule
			\multirow{3}{*}{Qwen3-8B} & \multirow{3}{*}{AIME 2025} & \textcolor{gray!65!black}{Baseline} & $0.571$ & $45.4$h & $43.8$h & $260$ & $251$ & $628$s & $61.0$ \\
		\cmidrule(lr){3-10}
			& & \textbf{Ours} & \textbf{0.679} & \textbf{36.9h} & \textbf{31.3h} & \textbf{190} & \textbf{161} & 699s & \textbf{77.7} \\
		\cmidrule(lr){3-10}
		& & \textcolor{blue!65!black}{Gain} & \textcolor{blue!65!black}{$+10.8$ pp} & \textcolor{blue!65!black}{$1.23\times$} & \textcolor{blue!65!black}{$1.40\times$} & \textcolor{blue!65!black}{$1.37\times$} & \textcolor{blue!65!black}{$1.56\times$} & \textcolor{blue!65!black}{$0.90\times$} & \textcolor{blue!65!black}{$1.27\times$} \\
		\midrule
			\multirow{3}{*}{Qwen3-4B} & \multirow{3}{*}{AIME 2025} & \textcolor{gray!65!black}{Baseline} & $0.590$ & $49.5$h & $45.4$h & $230$ & $211$ & $775$s & $66.0$ \\
		\cmidrule(lr){3-10}
			& & \textbf{Ours} & \textbf{0.662} & \textbf{29.5h} & \textbf{23.1h} & \textbf{140} & \textbf{110} & \textbf{757s} & \textbf{80.2} \\
		\cmidrule(lr){3-10}
		& & \textcolor{blue!65!black}{Gain} & \textcolor{blue!65!black}{$+7.3$ pp} & \textcolor{blue!65!black}{$1.68\times$} & \textcolor{blue!65!black}{$1.96\times$} & \textcolor{blue!65!black}{$1.64\times$} & \textcolor{blue!65!black}{$1.92\times$} & \textcolor{blue!65!black}{$1.02\times$} & \textcolor{blue!65!black}{$1.22\times$} \\
		\bottomrule
	\end{tabular*}
\end{table}

\paragraph{Matched baseline.}
Our main comparison is against the matched DeepSWE/GRPO++ recipe because Prefix Sampling is a drop-in controller with the same prompts, scalar rewards, $N=8$ rollout budget, and deployment protocol. Prefix/hint methods often add oracle prefixes or calibrated hints \citep{prefixrl2025,pope2025,stephint2025,adhint2025,selfhinting2025,hintgrpo2025}; SFT-RL hybrids add supervised or off-policy buffers \citep{luffy2025,srft2025,relift2025,uft2025}; and dynamic-allocation methods alter replacement sampling or rollout budgets \citep{yu2025dapo,dynamo2025,compositionrl2025}. We therefore evaluate PS as a fixed-rollout-budget controller over a matched GRPO++ backbone, not as a universal state-of-the-art comparison against protocol-changing methods. Appendix~\ref{app:method-comparison} summarizes these resource/protocol differences.

\subsection{Evaluation metrics}
The headline outcome metric is task-level Pass@1, evaluated on SWE-bench Verified for agentic RL and AIME 2025 for mathematical reasoning \citep{jimenez2024swebench,swebench_verified2024,aime2025}. All benchmark scores are averaged over 8 evaluation runs at each checkpoint. Efficiency is tracked by same-score convergence step, average wall-clock time per step, and valid rollout groups per batch, where a valid group is a non-degenerate group that survives group filtering. Outcome and efficiency metrics are reported across all four backbones where available. The mechanism analysis uses the 4B math run and reports distance to the $4/8$ target, parent-to-rerollout pass-count transitions, bucket-level correction toward $0.5$, and rollout-budget allocation across degenerate, skewed, and target-band outcomes.

\section{Results}
\label{sec:results}
Prefix Sampling is designed to steer skewed rollout groups toward the $50\%$ operating point. We first evaluate whether that design improves benchmark performance and training efficiency across all four backbones.

\subsection{Main benchmark performance}
\begin{figure}[!htbp]
	\centering
	\begin{minipage}[b]{0.48\linewidth}
		\centering
		\includegraphics[width=\linewidth]{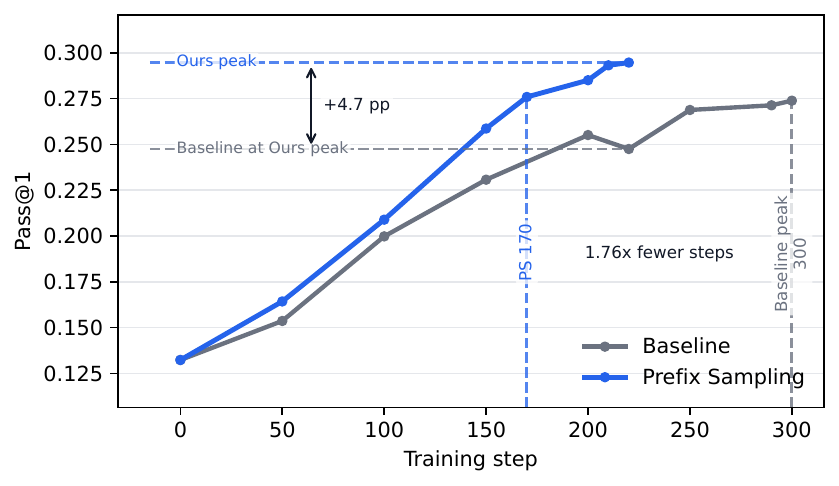}
		\subcaption{Qwen3-14B, SWE-bench Verified}
		\label{fig:swe-curve-14b}
	\end{minipage}
	\hfill
	\begin{minipage}[b]{0.48\linewidth}
		\centering
		\includegraphics[width=\linewidth]{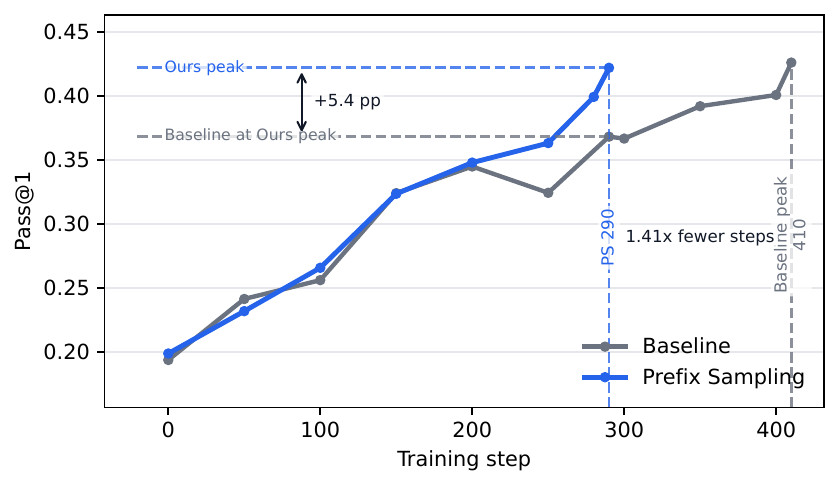}
		\subcaption{Qwen3-32B, SWE-bench Verified}
		\label{fig:swe-curve-32b}
	\end{minipage}
	\vspace{0.5em}
	\begin{minipage}[b]{0.48\linewidth}
		\centering
		\includegraphics[width=\linewidth]{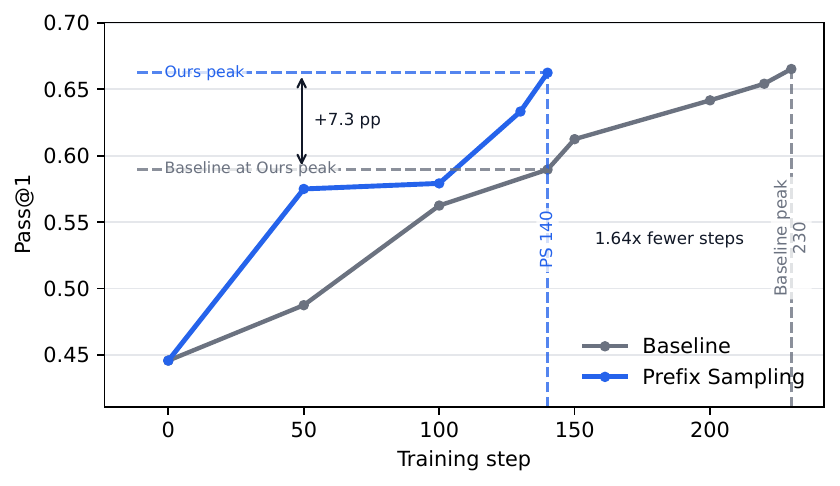}
		\subcaption{Qwen3-4B, AIME 2025}
		\label{fig:aime-curve-4b}
	\end{minipage}
	\hfill
	\begin{minipage}[b]{0.48\linewidth}
		\centering
		\includegraphics[width=\linewidth]{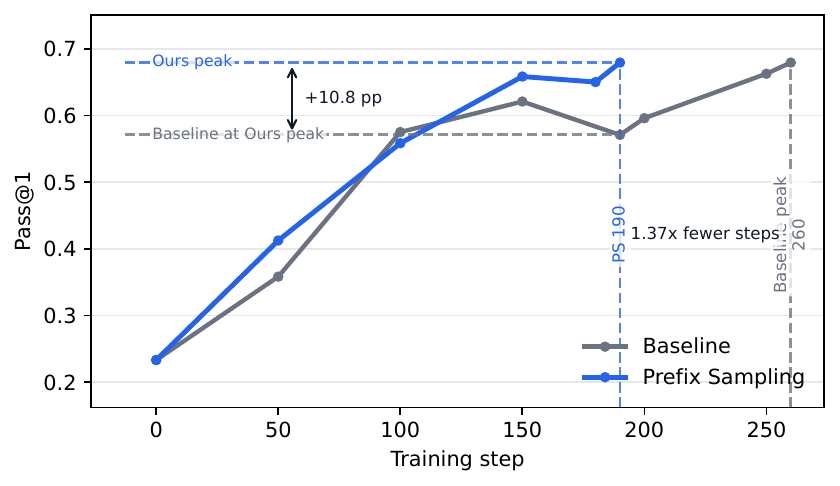}
		\subcaption{Qwen3-8B, AIME 2025}
		\label{fig:aime-curve-8b}
	\end{minipage}
\caption{\textbf{Benchmark performance over training} for Prefix Sampling and the baseline across agentic SWE-bench Verified runs and single-turn AIME 2025 runs. All points average 8 evaluation runs, and each panel is cropped to the checkpoint window used in Table~\ref{tab:main-results-summary}. Dashed vertical projections compare the baseline peak step with the earliest Prefix Sampling step reaching the same score level, or the same level within observed variability across the 8 evaluation runs where noted; dashed horizontal projections mark the same-step score gap at the Ours peak.}
	\label{fig:swe-curves}
\end{figure}

Table~\ref{tab:main-results-summary} separates two comparisons. The Pass@1 column is a same-checkpoint comparison at the Ours peak; the Eval/Train step and E2E columns compare convergence to the baseline reference level, allowing observed evaluation variability where noted in Figure~\ref{fig:swe-curves}. Entries are average-over-8 checkpoint evaluations, not independent training-seed estimates; Appendix~\ref{app:benchmark-eval-variability} reports the same average-over-8 means with variability estimates. Appendix~\ref{app:training-dynamics} plots the corresponding signal and system dynamics.

On SWE-bench Verified, Prefix Sampling reaches the baseline's high-score regime substantially earlier while giving a higher same-step score at the Ours peak. Qwen3-14B reaches the baseline peak level at step $170$ instead of the baseline's step $300$ and peaks at $0.295$, a $4.7$ pp same-step gain over the baseline at the Ours peak. Qwen3-32B reaches the baseline peak level within observed variability across the 8 evaluation runs at step $290$ instead of $410$ and reports a peak of $0.422$, a $5.4$ pp gain over the baseline at the same step, not over the baseline peak. Figure~\ref{fig:swe-curves} shows the same benchmark trajectories.

The mathematical-reasoning runs follow the same pattern on AIME 2025. Qwen3-4B reaches the baseline peak level within observed variability across the 8 evaluation runs at step $140$ instead of $230$ and shows a $7.3$ pp same-step score gap at the Ours peak. Qwen3-8B reaches the baseline peak level at step $190$ instead of $260$ and shows a $10.8$ pp same-step score gap. These results show that the evaluation-side convergence gain is not limited to stateful agent environments.

\subsection{Training efficiency and wall-clock cost}
The training-side columns of Table~\ref{tab:main-results-summary} explain where these gains come from. Prefix Sampling reaches the baseline training score on ordinary, non-replayed tasks earlier on all four backbones: $110$ versus $211$ steps on 4B, $161$ versus $251$ on 8B, $206$ versus $312$ on 14B, and $282$ versus $395$ on 32B. Valid groups also rise on every backbone: $25.9\!\to\!36.7$ on 14B, $30.9\!\to\!39.8$ on 32B, $66.0\!\to\!80.2$ on 4B, and $61.0\!\to\!77.7$ on 8B.

An independent system-level saving appears in the agentic runs, where prefix replay reconstructs the prefix state by executing saved responses and skips model generation for prefix tokens. Average per-step wall-clock drops from $1597$s to $1405$s on 14B and from $2353$s to $2146$s on 32B, equivalent to $1.14\times$ and $1.10\times$ step-time speedups. Combined with evaluation-side step reductions, this gives end-to-end speedups of $2.01\times$ on 14B and $1.55\times$ on 32B.

For single-turn mathematical reasoning, step-time is only diagnostic because these runs do not exercise replay-through-execution state reconstruction; the main mathematical-reasoning evidence is benchmark convergence and valid groups.

\subsection{Ablations}
\label{sec:ablations}
We isolate three design choices on the 4B AceReason-Math-Subset setup: whether prefixes from skewed groups are replayed at all, whether the replay length is adapted from bucket feedback, and whether the controller acts on both hard and easy buckets. Table~\ref{tab:ablation-summary} defines the four arms. \emph{PS-fix} uses replay with a fixed ratio $r=0.5$ for all four skewed buckets; \emph{PS-ada hard-only} keeps the adaptive controller but disables the easy-side $6/8$ and $7/8$ buckets; \emph{PS-ada} is the full four-bucket controller.

\begin{table}[t]
	\centering
	\setlength{\tabcolsep}{5pt}
		\caption{\textbf{4B AceReason-Math-Subset ablation summary.} Convergence is the peak step of the ordinary, non-replayed training-score trajectory marked in Figure~\ref{fig:ablation-summary}; valid groups/step is the mean number of non-degenerate rollout groups left after group filtering over each method's own pre-convergence window.}
	\label{tab:ablation-summary}
	\begin{tabular}{lcccccc}
		\toprule
			Method & Replay & Adaptive & Buckets & Conv. step $\downarrow$ & Speedup $\uparrow$ & \shortstack{Valid groups\\per step $\uparrow$} \\
			\midrule
			Baseline & No & No & --- & 211 & $1.00\times$ & 66.0 \\
			PS-fix & Yes & No & all four & 131 & $1.61\times$ & 85.1 \\
			PS-ada hard-only & Yes & Yes & hard only & 186 & $1.13\times$ & 71.9 \\
			PS-ada & Yes & Yes & all four & \textbf{110} & $\mathbf{1.92\times}$ & 80.2 \\
		\bottomrule
	\end{tabular}
\end{table}

\begin{figure}[!t]
	\centering
	\includegraphics[width=0.98\linewidth]{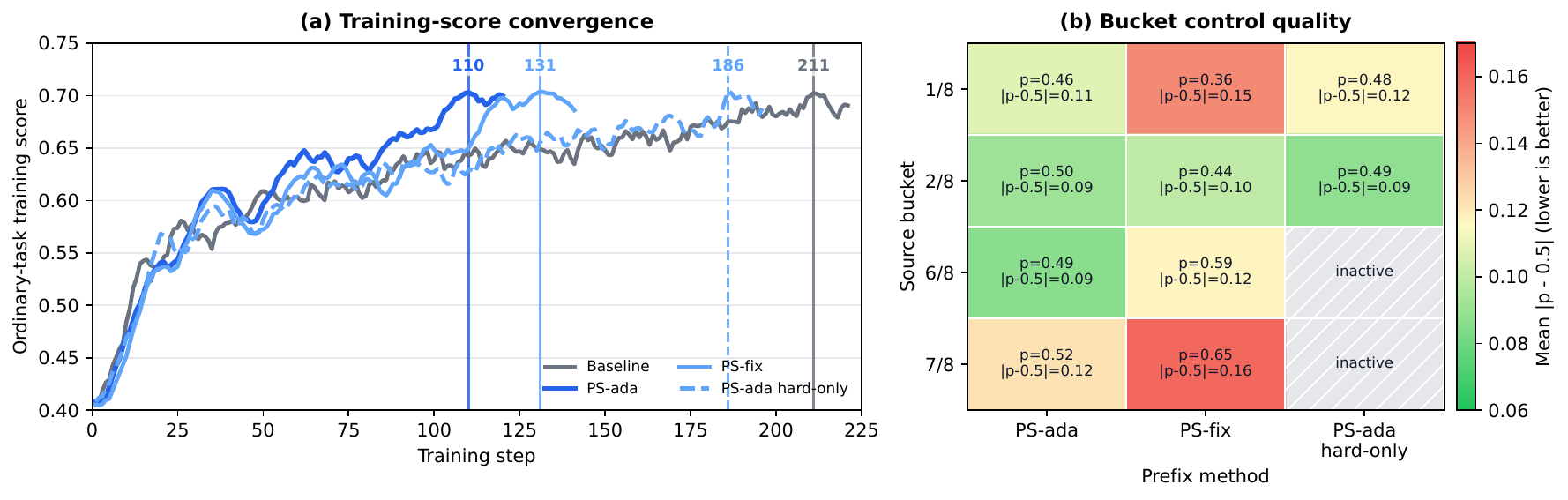}
	\caption{\textbf{4B AceReason-Math-Subset ablations.} Left: training-score trajectories on ordinary, non-replayed tasks, with vertical markers at each method's convergence step. Right: bucket-level control quality for the prefix-based arms, reporting mean rerollout pass rate $p$ and mean absolute distance $|p-0.5|$ before each method's convergence step; greener cells are closer to the $50\%$ target and gray cells are disabled by design.}
	\label{fig:ablation-summary}
\end{figure}

The left panel of Figure~\ref{fig:ablation-summary} shows the main outcome: the full PS-ada controller converges earliest at step $110$, followed by PS-fix at $131$, PS-ada hard-only at $186$, and the baseline at $211$. Enabling replay gives the largest jump: PS-fix reduces convergence from $211$ to $131$ steps, a $1.61\times$ speedup, by converting skewed groups into extra continuation-level training opportunities.

More valid groups help but do not fully explain the ordering. PS-fix keeps the largest number of valid groups per step, $85.1$, yet converges later than PS-ada, which keeps $80.2$. PS-ada hard-only keeps only $71.9$, closer to the baseline's $66.0$, and also converges much later. Thus the full method is not simply winning by retaining more groups after filtering; the rerollout groups also need to reach a useful operating regime.

The right panel explains the remaining gap. PS-ada keeps all four enabled buckets near the target, with mean rerollout pass rates $0.46$, $0.50$, $0.49$, and $0.52$ for the $1/8$, $2/8$, $6/8$, and $7/8$ buckets. PS-fix uses the same four source buckets but has no feedback loop, so the hard-side $1/8$ bucket stays too difficult at $p=0.36$ while the easy-side $7/8$ bucket drifts to $p=0.65$. PS-ada hard-only adapts the hard side successfully, with $p=0.48$ and $0.49$ for $1/8$ and $2/8$, but disables the easy side entirely. Thus adaptive control keeps continuations near $50\%$, while bidirectional coverage corrects both mostly failing and mostly passing groups.

Together, the ablations support the intended decomposition. Replay supplies most of the raw convergence reduction, adaptive control makes rerollout groups land closer to the high-signal regime, and hard/easy bidirectionality prevents the controller from ignoring the already-mastered side of the distribution. Only the combination yields the full $1.92\times$ speedup over the baseline.

\section{Analysis}
\label{sec:analysis}
The cross-scale results above establish the main empirical effect. We now use the 4B AceReason-Math-Subset run, where pass counts, parent-child mappings, and prefix metadata are recorded, to ask a narrower mechanism question: does Prefix Sampling move the controlled rerollout cohort toward the $50\%$ operating point? These diagnostics explain the results; they are not a substitute for the four-backbone evaluation above.

\subsection{Controlled rerollouts move toward the 50\% operating point}
Each sampled task group has $N=8$ rollouts, so a group with $k$ successful rollouts has pass rate $k/8$. We measure closeness to the target by the pass-count distance $|k-4|$, where $0$ is exactly balanced, $1$ is the $3/8$--$5/8$ target band, and $4$ is the degenerate $0/8$ or $8/8$ case.

\begin{figure}[!t]
	\centering
	\includegraphics[width=0.98\linewidth]{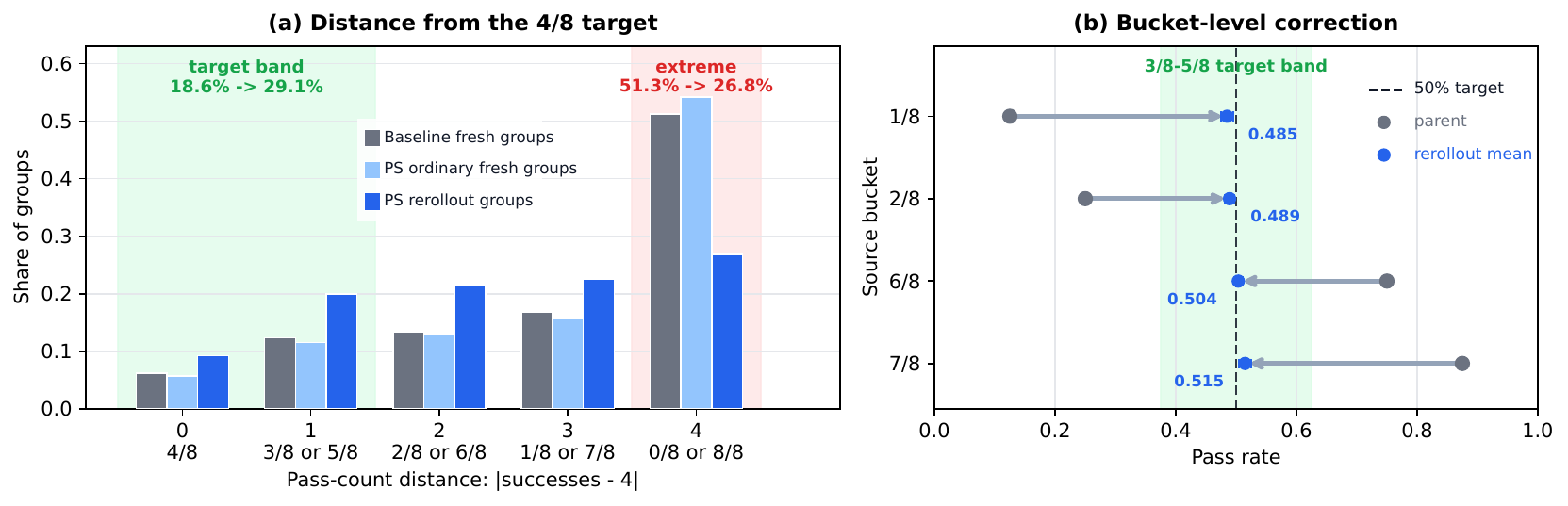}
		\caption{\textbf{Prefix Sampling moves controlled rerollouts toward the $50\%$ operating point on the 4B math run.} Left: pass-count distance from the $4/8$ target for baseline fresh groups, ordinary, non-replayed fresh groups from the PS run, and PS rerollout groups. Right: source-bucket pass rates before replay and mean rerollout pass rates after replay.}
	\label{fig:analysis-50pct-mechanism}
\end{figure}

The left panel of Figure~\ref{fig:analysis-50pct-mechanism} gives the aggregate distributional check. Prefix Sampling rerollouts reduce degenerate $0/8$ or $8/8$ mass from $51.3\%$ to $26.8\%$, increase the $3/8$--$5/8$ target band from $18.6\%$ to $29.1\%$, raise the exact $4/8$ bin from $6.2\%$ to $9.2\%$, and lower mean distance from $2.94$ to $2.38$. Ordinary, non-replayed fresh groups from the same Prefix Sampling run remain baseline-like, with target-band share $17.2\%$, extreme share $54.2\%$, and mean distance $3.01$, so the movement is localized to the controlled rerollout cohort. This gap also persists over training: rerollouts have a higher target-band share in $345$ of $372$ paired steps and a lower mean distance in $362$ of $372$.

The right panel checks that the aggregate shift is not hiding opposite bucket-level behavior. Parent pass rates remain $0.125$, $0.25$, $0.75$, and $0.875$ for the $1/8$, $2/8$, $6/8$, and $7/8$ buckets; after replay, their rerollout means are $0.485$, $0.489$, $0.504$, and $0.515$. Hard buckets are pulled upward by successful prefixes, easy buckets are pulled downward by failing prefixes, and all four controlled buckets end within $\pm 1.5$ percentage points of $0.5$.

\subsection{From pass-rate control to trainable signal}
\label{sec:budget-realloc}

The pass-rate movement in Figure~\ref{fig:analysis-50pct-mechanism} matters because degenerate groups provide little binary contrast: $0/8$ has no successful trajectory, $8/8$ has no failure, and both are typically removed before the policy update. By reducing extreme mass and increasing target-band mass, Prefix Sampling turns more of the same rollout budget into update-bearing groups that can carry reward entropy, RLOO advantage energy, and success--failure contrastive pairs. This keeps the causal claim narrow: Prefix Sampling does not balance every task or globally reshape the ordinary, non-replayed fresh-task distribution; it changes continuations for skewed partially successful groups, moving the controlled cohort toward the high-signal regime in Section~\ref{sec:why-target-50}. It also changes what the model practices: hard-side replay scaffolds missing steps, while easy-side replay creates self-correction. Appendix~\ref{app:mechanism-diagnostics} audits transitions and controller dynamics (Figures~\ref{fig:app-parent-child}--\ref{fig:app-adaptive-controller}); Appendix~\ref{app:case-studies} gives concrete hard/easy traces.

\section{Related Work}
\label{sec:related-work}

Skewed pass rates are known to degrade binary-reward RLVR efficiency, but existing methods usually mitigate them by reshaping the training distribution or adding scaffolding rather than explicitly targeting a pass-rate regime \citep{shao2024deepseekmath,yu2025dapo}.

\paragraph{Guidance and prefix scaffolding.}
PrefixRL and POPE condition on successful or oracle prefixes and optimize only the continuation \citep{prefixrl2025,pope2025}. StepHint and hint-GRPO adapt hint granularity or hint usage for difficult samples, while ADHint and Self-Hinting use difficulty-aware or model-generated hints \citep{stephint2025,hintgrpo2025,adhint2025,selfhinting2025}. These methods mainly rescue low-pass-rate tasks through offline traces, oracle/reference solutions, teacher chains, or hints; they do not address high-pass-rate groups with a symmetric handicapping intervention.

\paragraph{Distribution and protocol reshaping.}
DAPO's dynamic sampling filters all-fail/all-pass groups and samples replacements until the batch is filled, while DynaMO reallocates rollout budgets toward informative prompts \citep{yu2025dapo,dynamo2025}. LUFFY, SRFT, UFT, ReLIFT, Composition-RL, iGRPO, RLTF, and Experiential RL preserve signal through off-policy guidance, supervised mixtures, composite prompts, or richer interaction \citep{luffy2025,srft2025,uft2025,relift2025,compositionrl2025,igrpo2025,rltf2024,experientialrl2025}. These approaches are valuable, but they change batch construction, supervision, interaction, or rollout allocation. Prefix Sampling instead keeps the same rollout budget and changes the continuation states of retained skewed groups.

\paragraph{Targeting an informative regime.}
The closest conceptual line is SEELE, which uses item-response-style hint calibration to target balanced success--failure behavior \citep{seele2025}. This aligns with our motivation, but through hint calibration rather than binary-reward signal criteria. Our 50\% target is derived from reward entropy, RLOO advantage energy, and success--failure pair count, and the controller acts bidirectionally: successful prefixes help hard groups, while failing prefixes handicap easy groups. Appendix~\ref{app:method-comparison} and Table~\ref{tab:related-method-comparison} provide a detailed axis-by-axis comparison, including the agentic replay setting where state reconstruction is required rather than simple text concatenation.

\section{Conclusion}
\label{sec:conclusion}

For grouped binary-reward RLVR updates, reward-side signal is strongest near $p=0.5$ under the criteria studied here: reward entropy, group-filtering survival, RLOO advantage energy, and success--failure pair count. Prefix Sampling uses replay-through-execution and prefix masking to steer skewed rollout groups toward that regime, reaching the baseline high-score regime earlier across SWE-bench Verified and AIME 2025 and delivering $2.01\times$/$1.55\times$ end-to-end speedups on the stateful 14B/32B runs. Ablations attribute the gains to replay, bidirectional coverage, and adaptive control.

\renewcommand{\acksection}{\section*{Acknowledgments}}
\begin{ack}
We thank Mingzhe Lu from the University of Chinese Academy of Sciences for his valuable advice on refining the paper presentation and structure.
\end{ack}

\newpage
\bibliographystyle{unsrtnat}
\bibliography{references}

\newpage
\appendix

\section{Limitations and Scope}
\label{app:limitations}

\subsection{Scope of Claims}
Our claims are scoped to binary-reward RLVR with grouped rollouts, and all main experiments use $N=8$ rollouts per task. The largest-scale experiments target the intended stateful-agent setting: SWE-bench-style training. The finest-grained mechanism audits use the 4B mathematical-reasoning run where pass counts, prefix metadata, and parent-to-rerollout transitions are easiest to inspect. The mathematical-reasoning runs use the curated AceReason-Math-Subset described in Section~\ref{sec:experimental-recipe}, and the main comparison is a matched GRPO++ baseline under the same data and rollout budget. Protocol-changing alternatives are discussed as related work and in Appendix~\ref{app:method-comparison}.

\subsection{Agentic Replay Assumptions and Costs}
Prefix Sampling assumes that replay prefixes can be saved. In agentic environments, prefix replay does not introduce a separate sandbox restoration mechanism: during prefix steps, saved model responses are fed through the same action parser and environment executor used for ordinary model-generated responses, so the environment state is reconstructed by replaying the prefix trajectory through the standard rollout path. We did not observe prefix-replay-specific restoration failures in our runs, but framework-specific nondeterministic tools remain implementation-dependent.

The reported agentic step-time measurements are end-to-end measurements of this replay-through-execution training path. Prefix steps skip model generation and instead execute saved responses through the same environment path, so replay does not add a separate restore-latency component beyond ordinary environment execution. Prefix replay stores text prefixes rather than full sandbox snapshots; Table~\ref{tab:prefix-pool-memory} gives a conservative upper-bound estimate for this text prefix pool under our run configurations. The $50\%$ target, derived in Appendix~\ref{app:optimality}, concerns the binary-reward signal in grouped rollouts, while the controller hyperparameters are fixed to the settings in Table~\ref{tab:ps-config}.

\begin{table}[!htbp]
  \caption{\textbf{Conservative prefix-pool memory estimate.} We estimate the maximum text-prefix state as $\text{batch size}\times(\text{max prompt} + 0.95\times\text{max response})$ token IDs, with 4 bytes per int32 token ID. The estimate intentionally ignores compression, shorter realized prefixes, and CPU/off-GPU placement, so it is an upper bound for the text prefix pool rather than a measured runtime allocation.}
  \label{tab:prefix-pool-memory}
  \centering
  \setlength{\tabcolsep}{5pt}
  \renewcommand{\arraystretch}{1.08}
  \begin{tabular}{@{}lrrrr@{}}
    \toprule
    Setup & Batch & Max prompt & Max response & Upper-bound pool \\
    \midrule
    4B/8B math & 128 & 2048 & 16384 & 8.6 MiB \\
    14B SWE & 64 & 4096 & 32768 & 8.6 MiB \\
    32B SWE & 64 & 4096 & 65536 & 16.2 MiB \\
    \bottomrule
  \end{tabular}
\end{table}

\subsection{Supplemental Artifacts and Broader Impacts}
The anonymized supplemental material includes code, data, and reproduction instructions for the 4B/8B mathematical-reasoning experiments. This release covers the task-agnostic Prefix Sampling logic: pass-count bucket routing, prefix storage, adaptive prefix-ratio control, rerollout construction, and prefix masking. The software-engineering implementation uses the same controller and masking logic, with an environment adapter that replays saved agent responses through the action parser and executor to reconstruct tool, code, and conversation state. For the 14B/32B agentic software-engineering runs, the paper provides configurations, metrics, evaluation protocol, and training dynamics. Broader impacts include reduced rollout compute and energy use for binary-reward RLVR, while these efficiency gains could make stronger coding or reasoning agents cheaper to train. We release no trained models or deployed agents.

\section{Detailed Comparison with Prior Methods}
\label{app:method-comparison}

Prefix Sampling bridges the gaps identified in Section~\ref{sec:related-work} along four axes, summarized in Table~\ref{tab:related-method-comparison}. According to our reading of the released algorithms, it uses self-generated prefixes rather than oracle prefixes or external hints; performs bidirectional control over low- and high-pass-rate rollout groups; explicitly targets the $50\%$ operating point derived from binary-reward theory; and implements replay-through-execution state reconstruction for agentic reinforcement learning rather than text-only prefix concatenation. Prior methods satisfy one or two of these properties, but Prefix Sampling combines them into a single pass-rate steering framework that explicitly steers the rollout operating point.
The table is intended to clarify resource and protocol differences rather than claim that all methods are directly comparable under a single benchmark. DAPO-style dynamic sampling and DynaMO-style rollout allocation are omitted from the method-comparison table because they change batch construction or rollout budgeting rather than the prefix intervention itself; Section~\ref{sec:experimental-recipe} discusses why these are not matched fixed-rollout-budget baselines.

\begin{table}[!ht]
  \centering
  \setlength{\tabcolsep}{4pt}
  \caption{Comparison of related prefix, hint, and difficulty-control methods along key design axes. \emph{On-pol.\ pref.} indicates whether the prefix is produced by the policy being trained rather than by an oracle, teacher, or offline solution. \emph{Pref.-mask} means replayed or provided prefix tokens are excluded from the RL loss, so credit is assigned only to the continuation. \emph{Adapt.\ pref.} denotes online adjustment of prefix length or hint strength based on training feedback. \emph{Bi-dir.\ pref.} denotes explicit steering of both low-pass-rate tasks with helpful prefixes and high-pass-rate tasks with handicapping prefixes. \emph{Agentic replay} requires continuing from a reconstructed environment, tool, code, and conversation state rather than only concatenating text. \cmark/\pmark/\xmark{} indicate present, partial, and absent support.}
  \label{tab:related-method-comparison}
  \begin{tabular}{lccccc}
    \toprule
    Method & On-pol.\ pref. & Pref.-mask & Adapt.\ pref. & Bi-dir.\ pref. & Agentic replay \\
    \midrule
    \textbf{Prefix Sampling (PS)} & \cmark & \cmark & \cmark & \cmark & \cmark \\
    PrefixRL \citep{prefixrl2025}    & \xmark & \cmark & \xmark & \xmark & \xmark \\
    POPE \citep{pope2025}            & \xmark & \cmark & \xmark & \xmark & \xmark \\
    StepHint \citep{stephint2025}    & \xmark & \pmark & \pmark & \xmark & \xmark \\
    ADHint \citep{adhint2025}        & \xmark & \pmark & \cmark & \xmark & \xmark \\
    Self-Hinting \citep{selfhinting2025} & \pmark & \cmark & \cmark & \xmark & \xmark \\
    hint-GRPO \citep{hintgrpo2025}   & \xmark & \pmark & \cmark & \xmark & \xmark \\
    SEELE \citep{seele2025}          & \xmark & \pmark & \cmark & \xmark & \xmark \\
    UFT \citep{uft2025}              & \xmark & \xmark & \xmark & \xmark & \xmark \\
    \bottomrule
  \end{tabular}
\end{table}

\section{Hyperparameters and Training Configuration}
\label{app:config}
Tables~\ref{tab:run-config}--\ref{tab:ps-config} separate three concerns: scale- and task-specific run budgets, the matched GRPO++ optimizer backbone, and the Prefix Sampling controller. The Baseline and Prefix Sampling runs share the first two; Prefix Sampling differs by adding the controller in Table~\ref{tab:ps-config}. The model sources are Qwen3 \citep{qwen32025,qwen3_4b_instruct_2507,qwen3_8b,qwen3_14b,qwen3_32b}; the training/evaluation data and benchmarks are AceReason-Math/AIME 2025 \citep{acereason2025,aime2025} and R2E-Gym/SWE-bench Verified \citep{r2egym2025,jimenez2024swebench,swebench_verified2024}. The mathematical-reasoning runs train on AceReason-Math-Subset, constructed by the 8-rollout Qwen3-4B-Instruct-2507 subset-curation procedure described in Section~\ref{sec:experimental-recipe}; the software-engineering runs train on R2E-Gym-Subset. All benchmark scores are average-over-8 (avg8) evaluation means. The training implementation is based on verl v0.5.x \citep{verl2025}. AIME 2025 evaluation is run with EvalScope \citep{evalscope2024}; SWE-bench Verified evaluation uses the same verl validation pipeline as training, with the validation data replaced by SWE-bench Verified. Table~\ref{tab:reported-compute} gives convergence-window device-hour accounting by GPU type, and Table~\ref{tab:asset-licenses} records upstream sources and license/terms notes.

\begin{table}[!htbp]
  \caption{\textbf{Run-level training configuration.} Columns are experiment setups, and rows list model, data, thinking mode, hardware, batch, rollout, context, temperature, and agent budget. Batch is the pre-filter training batch size, and $N$ is the number of rollouts sampled per task group.}
  \label{tab:run-config}
  \centering
  \setlength{\tabcolsep}{4pt}
  \renewcommand{\arraystretch}{1.08}
  \begin{tabular}{@{}>{\raggedright\arraybackslash}p{0.15\linewidth}>{\raggedright\arraybackslash}p{0.20\linewidth}>{\raggedright\arraybackslash}p{0.16\linewidth}>{\raggedright\arraybackslash}p{0.20\linewidth}>{\raggedright\arraybackslash}p{0.20\linewidth}@{}}
    \toprule
    Setting & 4B math & 8B math & 14B SWE & 32B SWE \\
    \midrule
    Model & Qwen3-4B-Instruct-2507 & Qwen3-8B & Qwen3-14B & Qwen3-32B \\
    Train data & AceReason-Math-Subset & AceReason-Math-Subset & R2E-Gym-Subset & R2E-Gym-Subset \\
    Benchmark & AIME 2025 & AIME 2025 & SWE-bench Verified & SWE-bench Verified \\
    Thinking & off & off & on & on \\
    Hardware & $2\times8$ A800 & $4\times8$ A800 & $4\times8$ B200 & $4\times8$ B200 \\
    Batch size & 128 & 128 & 64 & 64 \\
    Rollouts/task & $N=8$ & $N=8$ & $N=8$ & $N=8$ \\
    Max prompt & 2048 & 2048 & 4096 & 4096 \\
    Max response & 16384 & 16384 & 32768 & 65536 \\
    Temperature & 1.0 & 1.0 & 1.0 & 1.0 \\
    Agent budget & no agent & no agent & 50 steps, 1200s & 100 steps, 1200s \\
    \bottomrule
  \end{tabular}
\end{table}

\begin{table}[!htbp]
  \caption{\textbf{Convergence-window device-hour accounting.} Device-hours are computed from Table~\ref{tab:main-results-summary} as the convergence-window wall-clock hours used for the same-score comparisons multiplied by the number of GPUs in Table~\ref{tab:run-config}. Because A800 and B200 hours are not hardware-normalized, totals are separated by accelerator type: Eval view $20{,}490$ B200-hours plus $3{,}898$ A800-hours; Train view $20{,}643$ B200-hours plus $3{,}499$ A800-hours. Eval and Train are two accounting views of the same runs and should not be added together. Unreported pilot or failed runs are excluded from these convergence-window totals.}
  \label{tab:reported-compute}
  \centering
  \setlength{\tabcolsep}{4pt}
  \renewcommand{\arraystretch}{1.05}
  \begin{tabular}{@{}llccc@{}}
    \toprule
    Setup & Row & GPUs & Eval dev.-h & Train dev.-h \\
    \midrule
    32B SWE & Baseline & 32 B200 & $8576.0$ & $8262.4$ \\
    32B SWE & Ours & 32 B200 & $5532.8$ & $5379.2$ \\
    14B SWE & Baseline & 32 B200 & $4259.2$ & $4428.8$ \\
    14B SWE & Ours & 32 B200 & $2121.6$ & $2572.8$ \\
    8B math & Baseline & 32 A800 & $1452.8$ & $1401.6$ \\
    8B math & Ours & 32 A800 & $1180.8$ & $1001.6$ \\
    4B math & Baseline & 16 A800 & $792.0$ & $726.4$ \\
    4B math & Ours & 16 A800 & $472.0$ & $369.6$ \\
    \bottomrule
  \end{tabular}
\end{table}

\begin{table}[!htbp]
  \caption{\textbf{Existing assets, frameworks, and license/terms notes.} We cite the upstream sources used for models, data, benchmarks, and software. For assets whose dataset card or official page does not list an explicit open license, we follow upstream terms and do not redistribute the original benchmark content in this paper.}
  \label{tab:asset-licenses}
  \centering
  \setlength{\tabcolsep}{3pt}
  \renewcommand{\arraystretch}{1.06}
  \begin{tabular}{@{}>{\raggedright\arraybackslash}p{0.20\linewidth}>{\raggedright\arraybackslash}p{0.38\linewidth}>{\raggedright\arraybackslash}p{0.36\linewidth}@{}}
    \toprule
    Asset & Upstream source and use & License / terms note \\
    \midrule
    Qwen3 variants & Hugging Face / Qwen; 4B/8B/14B/32B model backbones & Apache-2.0 model license as listed upstream \\
    AceReason-Math & Hugging Face / NVIDIA; math training source, with a local subset derived by pass-rate-based curation & CC-BY-4.0 as listed upstream \\
    R2E-Gym-Subset & Hugging Face / R2E-Gym; SWE training source & Use follows upstream terms; dataset card does not list a separate license \\
    SWE-bench Verified & SWE-bench / Princeton-NLP; SWE evaluation benchmark & MIT license for SWE-bench code/evaluation harness; benchmark instances derive from upstream GitHub repositories/issues/PRs and follow upstream terms \\
    AIME 2025 & Mathematical Association of America; math evaluation benchmark & Official competition content; used for evaluation and not redistributed \\
    verl v0.5.x & GitHub / verl project; training and SWE validation framework & Apache-2.0 software license \\
    EvalScope & GitHub / ModelScope; AIME evaluation framework & Apache-2.0 software license \\
    \bottomrule
  \end{tabular}
\end{table}

Table~\ref{tab:grpo-config} gives the shared GRPO++ backbone \citep{deepswe2025,shao2024deepseekmath,schulman2017ppo,williams1992rloo,yu2025dapo}. This table is intentionally optimizer-focused: task scale, context length, and hardware are isolated in Table~\ref{tab:run-config}, while Prefix Sampling-specific control knobs are isolated in Table~\ref{tab:ps-config}.

\begin{table}[!htbp]
  \caption{\textbf{Shared GRPO++ backbone.} The same backbone is used for the Baseline and Prefix Sampling runs at each model scale.}
  \label{tab:grpo-config}
  \centering
  \setlength{\tabcolsep}{4pt}
  \renewcommand{\arraystretch}{1.08}
  \begin{tabular}{@{}p{0.20\linewidth}p{0.78\linewidth}@{}}
    \toprule
    Component & Setting \\
    \midrule
    Clip High & Upper clip bound $0.28$ \\
    No KL Loss & KL regularization disabled \\
    No Reward Std. & Reward standard-deviation normalization disabled \\
    Len. Norm. & Length-normalized surrogate loss \\
    Leave-One-Out & Leave-one-out (RLOO) advantage estimation \\
    Compact Filtering & Mask invalid max-context, timeout, or max-step trajectories when applicable \\
    No Entropy Loss & Entropy coefficient $0.0$ \\
    Actor LR & $1\!\times\!10^{-6}$ \\
    PPO mini-batch & 64 \\
    PPO micro/GPU & 1 \\
    \bottomrule
  \end{tabular}
\end{table}

Table~\ref{tab:ps-config} lists the Prefix Sampling controller settings used in the main runs. The controller acts only on skewed but non-degenerate buckets; all non-degenerate ordinary groups still train normally, while degenerate groups are handled by group filtering.

\begin{table}[!htbp]
  \caption{\textbf{Prefix Sampling controller configuration.} The controller adapts each skewed bucket independently toward the $0.5$ rollout pass-rate target. Prefix ratio is measured over environment-interaction steps for SWE agents and over assistant tokens for single-turn math.}
  \label{tab:ps-config}
  \centering
  \setlength{\tabcolsep}{4pt}
  \renewcommand{\arraystretch}{1.08}
  \begin{tabular}{@{}p{0.24\linewidth}p{0.34\linewidth}@{}}
    \toprule
    Item & Value \\
    \midrule
    Variant & 4-bucket PS-ada \\
    Controlled buckets & $1/8$, $2/8$, $6/8$, $7/8$ \\
    Target pass rate & $0.5$ \\
    Initial prefix ratio & $0.5$ for each bucket \\
    Ratio bounds & $[0.05, 0.95]$ \\
    Exponential-moving-average (EMA) alpha & $0.05$ \\
    Deadzone & $0.03$ \\
    Step size & $0.05$ \\
    Cooldown & 5 updates \\
    \bottomrule
  \end{tabular}
\end{table}

\section{Additional Training Dynamics}
\label{app:training-dynamics}

This appendix expands the aggregate entries in Table~\ref{tab:main-results-summary}. The figures use the same panel order as the main benchmark figure: Qwen3-14B, Qwen3-32B, Qwen3-4B, and Qwen3-8B. Thin traces are raw logged metric values, thick traces are EMA-smoothed with factor $0.80$, and vertical markers denote the baseline convergence step, the first Prefix Sampling step reaching the same training score on ordinary, non-replayed tasks, and, when different, the Prefix Sampling convergence step.

\begin{figure}[!htbp]
  \centering
  \includegraphics[width=0.99\linewidth]{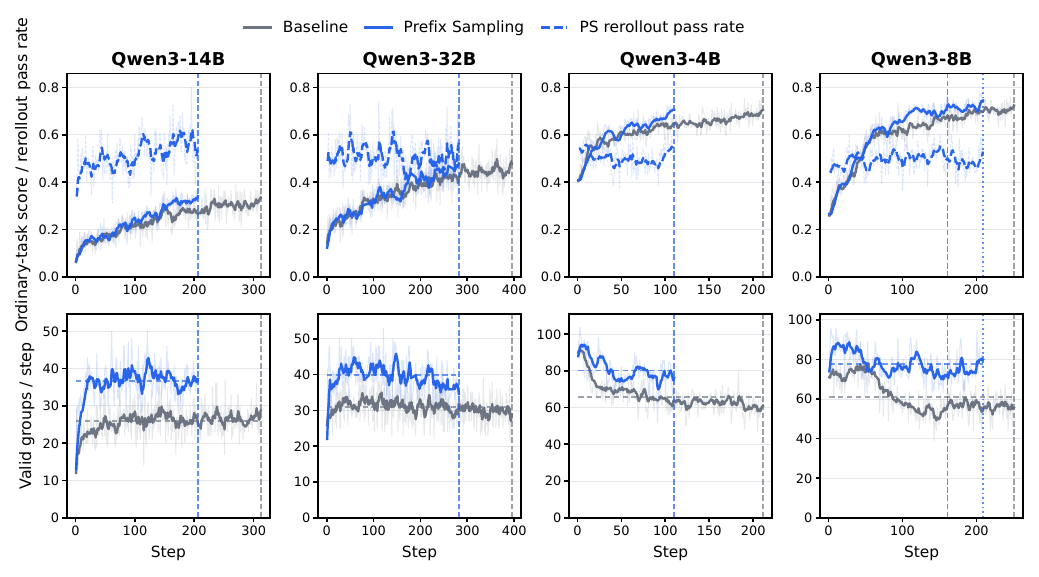}
  \caption{\textbf{Training-signal dynamics across all four backbones.} Top row: training score on ordinary, non-replayed tasks for the baseline and Prefix Sampling, plus the PS rerollout pass rate. Bottom row: valid rollout groups after group filtering. Gray curves are baselines; blue solid curves are Prefix Sampling training-score or per-step metrics; blue dashed curves are PS rerollout pass rates. Dashed horizontal segments in the row for valid groups mark each method's raw mean over its own convergence-cropped window.}
  \label{fig:app-training-dynamics-signal}
\end{figure}

Figure~\ref{fig:app-training-dynamics-signal} gives the training-signal view. Prefix Sampling reaches the baseline's training-score level on ordinary, non-replayed tasks earlier on all four backbones: step $206$ versus $312$ on 14B, $282$ versus $395$ on 32B, $110$ versus $211$ on 4B, and $161$ versus $251$ on 8B. At the same time, the PS rerollout pass-rate curves stay near the intended middle regime, with raw means $0.519$, $0.510$, $0.502$, and $0.497$ for 14B, 32B, 4B, and 8B. The row for valid groups explains the corresponding entries in Table~\ref{tab:main-results-summary}: prefix replay reduces the share of rerollout groups that collapse to all-fail or all-pass, increasing update-bearing groups and raising mean valid groups per step from $25.9$ to $36.7$ on 14B, $30.9$ to $39.8$ on 32B, $66.0$ to $80.2$ on 4B, and $61.0$ to $77.7$ on 8B.

\begin{figure}[!t]
  \centering
  \includegraphics[width=0.99\linewidth]{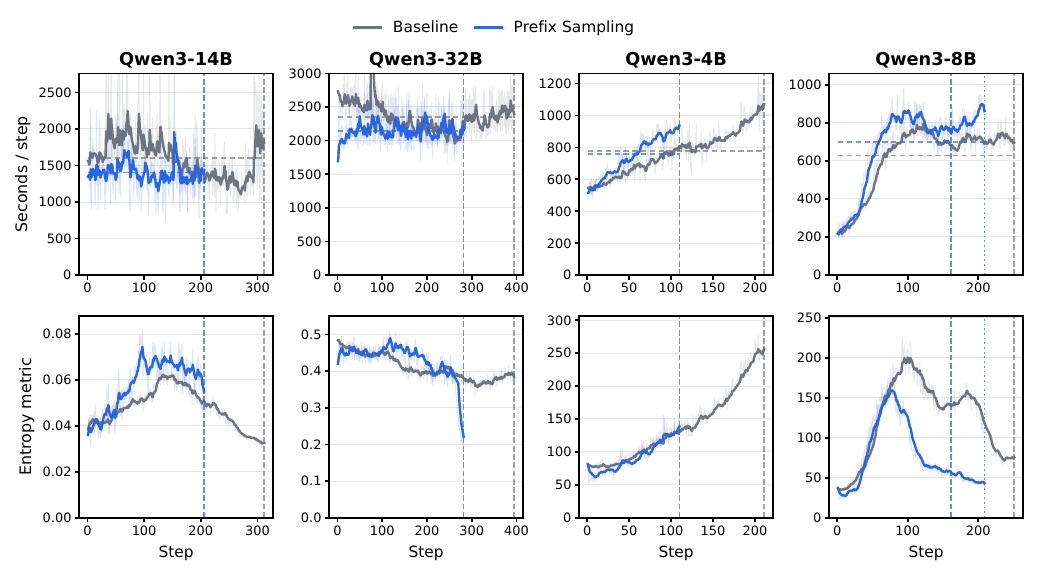}
  \caption{\textbf{System diagnostics across all four backbones.} Top row: wall-clock time per training step. Bottom row: logged policy entropy metric, distinct from the binary reward entropy in Appendix~\ref{app:optimality}. Gray curves are baselines and blue curves are Prefix Sampling. Dashed horizontal segments in the timing row mark each method's raw mean over its own convergence-cropped window. These diagnostics support the wall-clock claims on the stateful 14B/32B SWE-bench-style runs evaluated on SWE-bench Verified; on 4B/8B math, they are reported for completeness rather than used as the primary efficiency claim.}
  \label{fig:app-training-dynamics-system}
\end{figure}

Figure~\ref{fig:app-training-dynamics-system} reports system-side diagnostics. On the two stateful SWE-bench-style training runs evaluated on SWE-bench Verified, Prefix Sampling reduces mean wall-clock time per step from $1597$s to $1405$s on 14B and from $2353$s to $2146$s on 32B, while also maintaining higher logged policy entropy before convergence. The mathematical-reasoning runs do not exercise replay-through-execution agent-state reconstruction, so their step-time and entropy traces are diagnostics rather than the main claim: 4B step time is roughly comparable, while 8B is slower under Prefix Sampling. This is why the main 4B/8B evidence in the paper emphasizes benchmark convergence, valid groups, and the operating-point analysis in Section~\ref{sec:analysis}.

\subsection{Benchmark evaluation variability}
\label{app:benchmark-eval-variability}

Table~\ref{tab:benchmark-eval-variability} reports the average-over-8 (avg8) benchmark means used in the main results together with descriptive variability estimates from repeated evaluation records available at the same checkpoint. The table is not an independent-training-seed analysis; each row summarizes a fixed checkpoint and reports the standard deviation of the recovered evaluation records.

\begin{table}[!htbp]
  \centering
  \setlength{\tabcolsep}{3pt}
  \renewcommand{\arraystretch}{1.06}
  \caption{\textbf{Average-over-8 (avg8) variability for benchmark checkpoints used in the main results.} Mean is the benchmark score used in Table~\ref{tab:main-results-summary} and Figure~\ref{fig:swe-curves}; Ours denotes Prefix Sampling with PS-ada. Std. is estimated from repeated evaluation records available at the same checkpoint. These rows describe evaluation-level variability and do not estimate independent training-seed variance.}
  \label{tab:benchmark-eval-variability}
  \begin{tabular*}{\linewidth}{@{\extracolsep{\fill}}lllcc@{}}
    \toprule
    Model & Checkpoint role & Step & Mean & Std. \\
    \midrule
    14B & Baseline at Ours peak & 220 & 0.247 & 0.008 \\
    14B & Ours peak & 220 & 0.295 & 0.009 \\
    14B & Baseline peak & 300 & 0.274 & 0.013 \\
    14B & Ours reaches baseline peak & 170 & 0.276 & 0.012 \\
    32B & Baseline at Ours peak & 290 & 0.368 & 0.005 \\
    32B & Ours peak / reaches baseline within variability & 290 & 0.422 & 0.011 \\
    32B & Baseline peak reference & 410 & 0.426 & 0.006 \\
    4B & Baseline at Ours peak & 140 & 0.590 & 0.056 \\
    4B & Ours peak / reaches baseline within variability & 140 & 0.662 & 0.055 \\
    4B & Baseline peak reference & 230 & 0.665 & 0.042 \\
    8B & Baseline at Ours peak & 190 & 0.571 & 0.051 \\
    8B & Ours peak / reaches baseline & 190 & 0.679 & 0.043 \\
    8B & Baseline peak reference & 260 & 0.679 & 0.035 \\
    \bottomrule
  \end{tabular*}
\end{table}

\section{Optimality of $p=0.5$ for Binary-Reward Learning Signal}
\label{app:optimality}

This appendix derives the sense in which a rollout pass rate of $p=0.5$ is the optimal operating point for binary-reward RL, using standard Bernoulli entropy and the Group Relative Policy Optimization (GRPO) backbone with leave-one-out (RLOO) advantages used in our training recipe \citep{cover2006information,shao2024deepseekmath,williams1992rloo}. The claim is not that a final trained model should solve each task with only $50\%$ probability. Rather, during training, a task group whose rollouts are balanced between success and failure carries the strongest binary learning signal. The derivation assumes a fixed task, $N$ independently sampled rollouts from the current policy, binary terminal rewards $r_i \in \{0,1\}$, and pass probability $p=\Pr(r_i=1)$. In an observed group, let $K=k$ be the number of successes and $\hat p=k/N$ be the empirical pass rate. Table~\ref{tab:summary-50} summarizes the complementary quantities below.

\begin{table}[!ht]
  \centering
  \caption{Complementary measures of binary-reward learning signal. Each measure is maximized at $p=0.5$, or, for observed groups with even $N$, at $k=N/2$.}
  \label{tab:summary-50}
  \begin{tabular}{llcc}
    \toprule
    Perspective & Quantity & Expression & Maximized at \\
    \midrule
    Information          & Reward entropy        & $H(p)$ & $p=0.5$ \\
    Filtering            & Non-degenerate group prob. & $1-p^N-(1-p)^N$ & $p=0.5$ \\
    Optimization         & RLOO advantage energy & $k(N-k)/(N-1)^2$ & $k=N/2$ \\
    Credit assignment    & Contrastive pairs     & $k(N-k)$ & $k=N/2$ \\
    Expected contrast    & Expected pair count   & $N(N-1)p(1-p)$ & $p=0.5$ \\
    \bottomrule
  \end{tabular}
\end{table}

\subsection{Reward Entropy}

With binary rewards, a single rollout outcome is a Bernoulli random variable with success probability $p$. Its information content is the entropy
\begin{equation}
  H(p) = -p \log_2 p - (1-p)\log_2(1-p).
\end{equation}
When $p$ approaches $0$ or $1$, the outcome becomes predictable and each additional rollout provides little new information. Taking derivatives,
\begin{equation}
  \frac{dH}{dp}
  = -\log_2 p - \frac{1}{\ln 2} + \log_2(1-p) + \frac{1}{\ln 2}
  = \log_2\!\left(\frac{1-p}{p}\right),
\end{equation}
so the only stationary point satisfies
\begin{equation}
  \log_2\!\left(\frac{1-p}{p}\right)=0
  \;\Longleftrightarrow\;
  1-p=p
  \;\Longleftrightarrow\;
  p=0.5.
\end{equation}
The second derivative is
\begin{equation}
  \frac{d^2H}{dp^2}
  = -\frac{1}{p\ln 2} - \frac{1}{(1-p)\ln 2} < 0
  \qquad \text{for all } p \in (0,1),
\end{equation}
so $H(p)$ is uniquely maximized at $p=0.5$, where $H(0.5)=1$ bit. For $N$ independent rollouts from the same task, the joint entropy of the reward vector is $N H(p)$, which has the same maximizer. Thus the conclusion is not an artifact of considering one rollout at a time.

The finite-$N$ numbers used in our experiments make the loss from skew concrete. For $N=8$, a group with empirical pass rate $\hat p=4/8$ has empirical Bernoulli entropy $H(\hat p)=1.00$ bit per rollout, a $2/8$ or $6/8$ group has $H(\hat p)\approx 0.81$ bits, and a $1/8$ or $7/8$ group has $H(\hat p)\approx 0.54$ bits. A $1/8$ group therefore carries slightly more than half the per-rollout information of a balanced group. Entropy measures information available in the reward; the following perspectives ask whether that information survives filtering and turns into optimization signal.

\subsection{Survival Under Group Filtering}

Our training recipe filters out all-fail and all-pass groups before the RL update, following the group-filtering logic used in GRPO-style RLVR systems and later formalized in DAPO's dynamic sampling \citep{shao2024deepseekmath,yu2025dapo}. For $K \sim \mathrm{Binomial}(N,p)$, the probability that a sampled group is non-degenerate is
\begin{equation}
  \Pr(0<K<N) = 1 - \Pr(K=0) - \Pr(K=N)
  = 1 - (1-p)^N - p^N.
\end{equation}
This quantity is maximized when the probability of the two degenerate extremes is minimized. Differentiating,
\begin{equation}
  \frac{d}{dp}\Pr(0<K<N)
  = N(1-p)^{N-1} - Np^{N-1}.
\end{equation}
The stationary point satisfies $(1-p)^{N-1}=p^{N-1}$, hence $p=0.5$. The second derivative is
\begin{equation}
  -N(N-1)\!\left((1-p)^{N-2}+p^{N-2}\right)<0,
\end{equation}
so this is the unique maximum on $(0,1)$. For $N=8$, the non-degenerate probability is $99.22\%$ at $p=0.5$, $89.99\%$ at $p=0.25$, and $65.64\%$ at $p=0.125$. Balanced groups are therefore not only more informative conditional on being used; they are also much less likely to be discarded as all-fail or all-pass.

\subsection{Mean-Centered and Leave-One-Out Advantage Signals}

The optimization view asks how much reward contrast becomes an advantage signal. With binary rewards and $k$ successes in a group, the empirical pass rate is $\hat p=k/N$. For ordinary mean-centered advantages,
\begin{align}
  A_i = 1-\hat p \quad (r_i=1), \qquad
  A_i = -\hat p \quad (r_i=0).
\end{align}
The within-group mean of $A$ is zero, and a direct calculation gives
\begin{equation}
  \mathrm{Var}_{\mathrm{centered}}(A)
  = \hat p(1-\hat p)^2 + (1-\hat p)\hat p^2
  = \hat p(1-\hat p).
\end{equation}
As a function of $\hat p$, this is a downward parabola maximized at $\hat p=0.5$.

Our experiments use leave-one-out (RLOO) advantage estimation \citep{williams1992rloo}, so we also spell out the exact finite-group form. The leave-one-out baseline for rollout $i$ is the average reward of the other $N-1$ rollouts. If $r_i=1$, the remaining group has $k-1$ successes, so
\begin{equation}
  A_i^{\mathrm{RLOO}}
  = 1 - \frac{k-1}{N-1}
  = \frac{N-k}{N-1}.
\end{equation}
If $r_i=0$, the remaining group has $k$ successes, so
\begin{equation}
  A_i^{\mathrm{RLOO}}
  = 0 - \frac{k}{N-1}
  = -\frac{k}{N-1}.
\end{equation}
The RLOO advantages take two outcome-dependent values with opposite signs; their magnitudes become equal only at $k=N/2$. Averaging squared RLOO advantages within the observed group yields
\begin{align}
  \frac{1}{N}\sum_{i=1}^{N}\left(A_i^{\mathrm{RLOO}}\right)^2
  &= \frac{1}{N}
     \left[
       k\left(\frac{N-k}{N-1}\right)^2
       +(N-k)\left(\frac{k}{N-1}\right)^2
    \right] \\
  &= \frac{k(N-k)}{(N-1)^2}.
\end{align}
The denominator is constant for fixed $N$, so RLOO advantage energy is maximized exactly when $k(N-k)$ is maximized, i.e., at $k=N/2$. For $N=8$, this value is $16/49$ at $k=4$, $12/49$ at $k=2$ or $k=6$, and $7/49$ at $k=1$ or $k=7$. Thus the same $50\%$ target follows under the leave-one-out advantage estimator used in the GRPO++ backbone.

This argument should be read as a statement about reward-side signal strength, not as a claim that the realized policy-gradient norm is determined only by $p$. The gradient also depends on the continuation trajectories and their log-probability derivatives. However, when the reward contrast collapses, no optimizer can extract a strong binary-reward update from that group; balancing the pass rate maximizes the part of the signal controlled by the rollout outcomes.

\subsection{Success--Failure Contrastive Pairs}

The credit-assignment view counts how many success--failure comparisons are available within a rollout group. With $N$ rollouts and $k$ successes, each successful trajectory can be compared against each failing trajectory, giving
\begin{equation}
  C(k) = k(N-k).
\end{equation}
Completing the square,
\begin{equation}
  C(k) = -\left(k-\frac{N}{2}\right)^2 + \frac{N^2}{4},
\end{equation}
so the maximum occurs at $k=N/2$ when $N$ is even, and at the two adjacent integers when $N$ is odd. For our $N=8$ setting, Table~\ref{tab:contrastive-pairs} shows the complete count.

\begin{table}[!ht]
  \centering
  \caption{Success--failure contrastive-pair count for the $N=8$ rollout groups used in our experiments. The exact optimum is $4/8$, while $3/8$ and $5/8$ preserve $94\%$ of the maximum contrastive-pair count.}
  \label{tab:contrastive-pairs}
  \begin{tabular}{cccc}
    \toprule
    Successes $k$ & Pass rate & $C(k)=k(8-k)$ & Relative to max. \\
    \midrule
    0 & $0/8$ & 0  & $0.00$ \\
    1 & $1/8$ & 7  & $0.44$ \\
    2 & $2/8$ & 12 & $0.75$ \\
    3 & $3/8$ & 15 & $0.94$ \\
    4 & $4/8$ & 16 & $1.00$ \\
    5 & $5/8$ & 15 & $0.94$ \\
    6 & $6/8$ & 12 & $0.75$ \\
    7 & $7/8$ & 7  & $0.44$ \\
    8 & $8/8$ & 0  & $0.00$ \\
    \bottomrule
  \end{tabular}
\end{table}

This table also explains why the empirical analysis often reports the $3/8$--$5/8$ band rather than only exact $4/8$. Finite rollout groups are noisy, and a controller that tries to force every rerollout group to exact $4/8$ would oscillate. The neighboring $3/8$ and $5/8$ groups remain almost maximally contrastive, whereas $1/8$ and $7/8$ retain less than half of the maximum pair count. The target band is therefore a stable practical proxy for the theoretical optimum.

The same conclusion holds in expectation before the group is sampled. If $K\sim\mathrm{Binomial}(N,p)$, then
\begin{align}
  \mathbb{E}[K(N-K)]
  &= N\,\mathbb{E}[K] - \mathbb{E}[K^2] \\
  &= N(Np) - \left(Np(1-p)+N^2p^2\right) \\
  &= N(N-1)p(1-p),
\end{align}
which is again maximized at $p=0.5$. Thus both the realized count $k(N-k)$ and its expectation under policy sampling identify the same operating point.

\subsection{Summary}

These derivations identify $p=0.5$ from several independent angles. It maximizes binary reward entropy, maximizes the probability that a group survives group filtering, maximizes mean-centered advantage variance and RLOO advantage energy, and maximizes both observed and expected success--failure contrastive pairs. Skewed non-degenerate groups such as $1/8$ or $7/8$ are not useless, but they provide much weaker binary learning signal than balanced groups. Prefix Sampling is designed around this fact: rather than merely drawing more rollouts, it changes the continuation state so that skewed but partially solved rollout groups move toward the high-signal region around $4/8$.
This is a statement about the binary-reward signal available to a grouped policy-gradient update, not a theorem that every global curriculum should keep all tasks at $50\%$ success or that trajectory diversity and policy-gradient geometry are irrelevant.

\section{Additional Mechanism Diagnostics}
\label{app:mechanism-diagnostics}

This appendix expands the 4B mechanism analysis in Section~\ref{sec:analysis}. Appendix~\ref{app:parent-child-transition} audits parent-to-rerollout transitions at the pass-count level, while Appendix~\ref{app:adaptive-controller-dynamics} checks whether the adaptive prefix ratios produce the expected closed-loop behavior.

\subsection{Parent-to-rerollout transition audit}
\label{app:parent-child-transition}

\begin{figure}[!ht]
  \centering
  \includegraphics[width=0.98\linewidth]{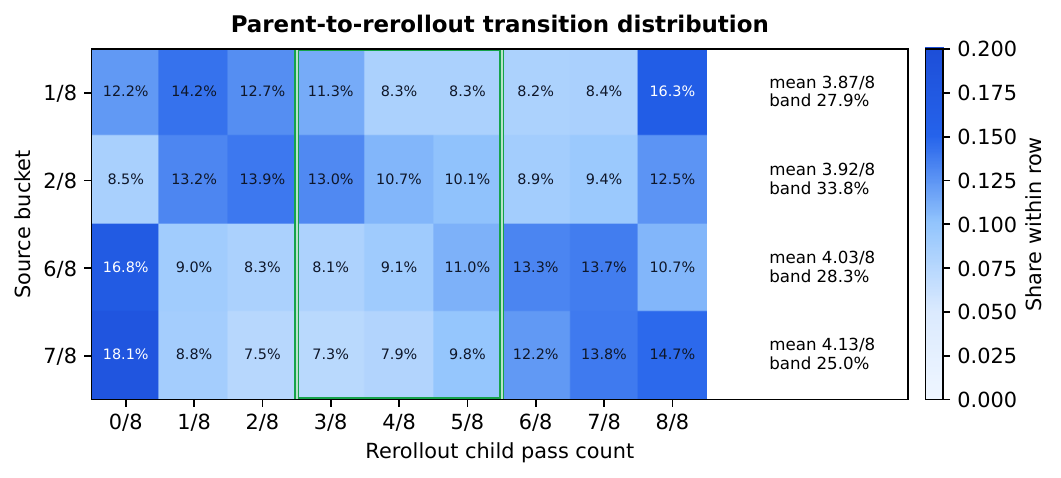}
  \caption{\textbf{Parent-to-rerollout transition distribution for the 4B adaptive controller.} Rerollout groups generated from hard-side parents shift upward, while those from easy-side parents shift downward. The green outline marks the $3/8$--$5/8$ target band, and the values on the right report the mean child pass count over all rerollout groups and target-band share for each source bucket.}
  \label{fig:app-parent-child}
\end{figure}

Figure~\ref{fig:app-parent-child} gives the transition audit behind the bucket-level correction summary in Figure~\ref{fig:analysis-50pct-mechanism}. Rows are conditioned on the source bucket that supplied the prefix, and columns show the full child rerollout pass-count distribution from $0/8$ through $8/8$. This view keeps the degenerate child rerollout groups in the table, so the reader can see both outcomes at once: some rerollout groups still collapse to all-fail or all-pass, but each source bucket also sends a visible mass into the $3/8$--$5/8$ target band around the $4/8$ operating point.

The hard-side rows move inward from below. A $1/8$ parent has a mean child pass count of $3.87/8$ and $27.9\%$ target-band mass; a $2/8$ parent has a mean child pass count of $3.92/8$ and $33.8\%$ target-band mass. The easy-side rows show the symmetric effect: $6/8$ and $7/8$ parents have mean child pass counts of $4.03/8$ and $4.13/8$, with $28.3\%$ and $25.0\%$ target-band mass. Because this audit keeps the residual $0/8$ and $8/8$ outcomes, it does not overstate the controller's precision. The direction is still clear: successful prefixes lift hard buckets, failing prefixes handicap easy buckets, and both sides are pulled toward the informative middle.

\subsection{Adaptive-prefix controller dynamics}
\label{app:adaptive-controller-dynamics}

\begin{figure}[!ht]
  \centering
  \includegraphics[width=0.98\linewidth]{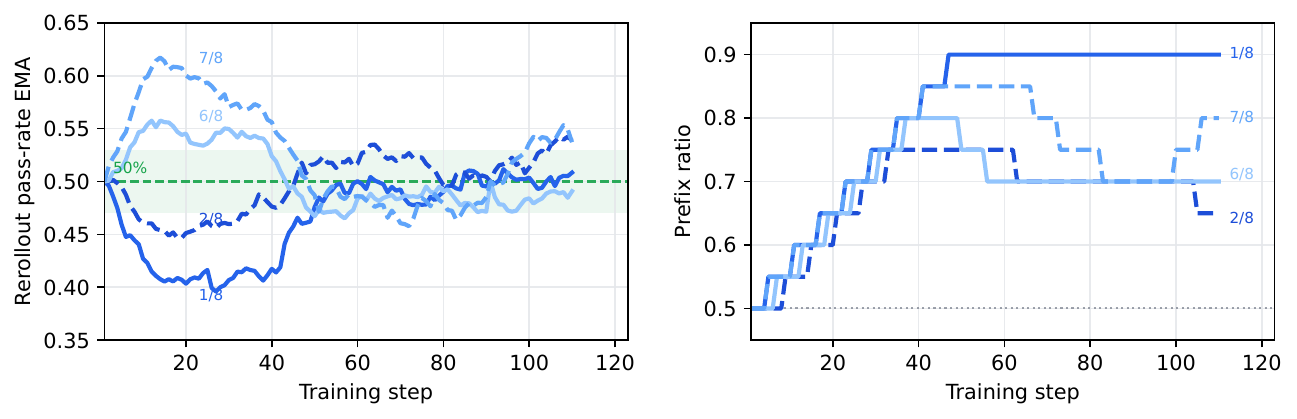}
  \caption{\textbf{Adaptive-controller dynamics on the 4B run.} Left: the rerollout pass-rate EMA used as bucket-level feedback. Right: the adaptive prefix ratio applied to each source bucket. The plotted window ends at the 4B Prefix Sampling convergence step.}
  \label{fig:app-adaptive-controller}
\end{figure}

Section~\ref{sec:select-control} defines the adaptive controller. Here we audit whether its observed time series match the intended behavior. Figure~\ref{fig:app-adaptive-controller} plots the bucket-level feedback EMAs and the corresponding prefix ratios on the 4B adaptive run through the convergence step used in Table~\ref{tab:main-results-summary}. The feedback traces start from the neutral initialization at $0.5$, temporarily separate as the controller observes skewed rerollouts, and return close to the target by convergence: the final EMA values are $0.508$, $0.539$, $0.491$, and $0.537$ for the $1/8$, $2/8$, $6/8$, and $7/8$ buckets.

The ratio traces show that this behavior is not produced by a single global replay length. By step $110$, the four bucket-specific ratios have separated to $0.90$, $0.65$, $0.70$, and $0.80$, indicating that different source buckets require different intervention strengths. The directions also match the controller's semantics. For too-hard buckets ($1/8$ and $2/8$), increasing the successful-prefix ratio gives the continuation a longer head start, and the feedback EMAs rise from the early under-target region back toward $0.5$. For too-easy buckets ($6/8$ and $7/8$), increasing the failing-prefix ratio commits the continuation to more of a failing trajectory, and the feedback EMAs fall from the early over-target region toward $0.5$. The important check is the joint pattern: despite non-uniform ratio schedules and opposite intervention semantics on the two sides, all four rerollout EMAs remain in a narrow band around $0.5$ by convergence. This is consistent with the expected closed-loop effect: bucket-specific prefix ratios compensate for bucket-specific drift and keep the controlled rerollout cohort near the intended operating point.

\section{Case Study Details}
\label{app:case-studies}
The two cases below illustrate the two directions of the Prefix Sampling intervention with one example each, both drawn from the 4B AceReason-Math-Subset run using Qwen3-4B-Instruct-2507 \citep{acereason2025,qwen3_4b_instruct_2507}. Each case is formatted as a compact audit panel followed by an evidence trail. The panel states the task, parent group, replayed prefix, and rerollout group; the evidence trail then separates model-generated snippets from our interpretation. Ellipses mark omitted text from longer trajectories.

\subsection{Hard-side rescue: missing overlap correction}
\noindent
\rule{\linewidth}{0.4pt}
\noindent\textbf{Problem.} Surface area of a $3 \times 3 \times 3$ wooden cube after three centered $1 \times 1$ tunnels are drilled through the three pairs of opposite faces.\par
\noindent\textbf{Gold answer.} $72$.\par
\noindent\textbf{Parent group.} $2/8$ correct; answer distribution $84{:}6$, $72{:}2$.\par
\noindent\textbf{Prefix.} Successful parent trajectory from the $2/8$ bucket; prefix length $4644$.\par
\noindent\textbf{Rerollout group.} $4/8$ correct; answer distribution $72{:}4$, $78{:}3$, $84{:}1$.
\par\noindent\rule{\linewidth}{0.4pt}

\noindent\textbf{Evidence trail.} The parent failures treat the three tunnels as independent and add all tunnel-wall area:
\begin{description}
\item[Parent failure, pred. 84.] ``Remaining outer surface: $48$ m\textsuperscript{2}. Internal tunnel surfaces: $36$ m\textsuperscript{2}. Total surface area $=48+36=84$ m\textsuperscript{2}.''
\item[Parent success, pred. 72.] ``The holes intersect at the center of the cube ... we must be careful not to double-count the internal surface areas, or to subtract overlapping areas where the hole walls meet.''
\item[Replayed prefix state.] ``The three holes cross each other at the center. So, we must be careful not to double-count the internal surface areas ... because the holes intersect, internal surfaces from different holes may overlap or share area.''
\item[Rerollout partial failure, pred. 78.] ``The three tunnels contribute $36$ m\textsuperscript{2} internally, but they intersect in three pairs ... remove $3 \times 2 = 6$ m\textsuperscript{2} from the $36$. So inner surface area $=36-6=30$ ... total surface area $=48+30=78$.''
\item[Rerollout success, pred. 72.] ``Each tunnel has 4 walls ... in the central $1 \times 1 \times 1$ cube ... subtract $3 \times 4 = 12$ m\textsuperscript{2}. So internal surface $=3 \times 12 - 12 = 24$. Final surface area $=54-6+24=72$.''
\end{description}
\noindent\textbf{Takeaway.} This case shows the hard-side role of Prefix Sampling. A rare policy-generated successful trajectory becomes an on-policy scaffold, and the fresh continuations produce contrast around the exact reasoning step that the parent group usually misses: whether the central overlap removes $0$, $6$, or $12$ square meters of tunnel-wall area.

\subsection{Easy-side handicap: recovery from a common misconception}
\noindent
\rule{\linewidth}{0.4pt}
\noindent\textbf{Problem.} Largest number of students, out of $50$, who can score strictly above the class mean.\par
\noindent\textbf{Gold answer.} $49$.\par
\noindent\textbf{Parent group.} $7/8$ correct; answer distribution $49{:}7$, $25{:}1$.\par
\noindent\textbf{Prefix.} Failing parent trajectory from the $7/8$ bucket; prefix length $273$.\par
\noindent\textbf{Rerollout group.} $4/8$ correct; answer distribution $25{:}4$, $49{:}4$.
\par\noindent\rule{\linewidth}{0.4pt}

\noindent\textbf{Evidence trail.} The parent group is mostly correct, but the single failure exposes the tempting misconception that ``above average'' means at most half the class:
\begin{description}
\item[Parent success, pred. 49.] ``49 students score 100. 1 student scores 0. Total sum $=4900$. Mean $=98$. Now, the 49 students who got 100 are above 98, so they are above the mean.''
\item[Parent failure, pred. 25.] ``By definition, half of the scores are above the mean and half below ... it is impossible for more than half of the students to score above the mean, because the mean is the balance point of the data.''
\item[Replayed prefix state.] ``The mean is the sum of all scores divided by the number of students. By definition, half of the scores are above the mean and half below ... it is impossible for more than half of the students to score above the mean.''
\item[Rerollout failure, pred. 25.] ``You cannot have more than 25 students strictly above the mean ... the maximum number of students who can score strictly above the mean is 25.''
\item[Rerollout success, pred. 49.] ``\textbf{Actually, no. There is no rule that says only half the students can score above the mean} ... 49 students scored 1, 1 student scored 0. Mean $=49/50=0.98$. Each of the 49 students scored $1>0.98$.''
\end{description}
\noindent\textbf{Takeaway.} This case shows the easy-side role of Prefix Sampling. A mostly solved task would otherwise provide little new signal, because most parent rollouts already repeat the answer $49$. Replaying the model's own failing branch turns the task into a self-correction exercise: the continuation first inherits the plausible but false ``half above the mean'' heuristic, then explicitly rejects it in the highlighted sentence and constructs a counterexample. Thus the easy-side intervention does more than make an easy problem harder; it creates direct training signal for detecting and repairing a tempting wrong reasoning branch.

\end{document}